\documentclass{article}
\usepackage{spconf,amsmath,graphicx,hyperref}
\usepackage{amsmath}
\usepackage{amssymb}
\usepackage{booktabs}
\usepackage{multirow}
\usepackage{fancyhdr}

\title{Latent Domain Prompt Learning for Vision-Language Models}
\name{Zhixing Li \qquad Arsham Gholamzadeh Khoee \qquad Yinan Yu\thanks{We acknowledge the National Academic Infrastructure for Supercomputing in Sweden (NAISS), partially funded by the Swedish Research Council through grant agreement no. 2022-06725, for awarding this project access to the LUMI supercomputer, owned by the EuroHPC Joint Undertaking and hosted by CSC (Finland) and the LUMI consortium.}}
  
  \address{Chalmers University of Technology\\
    Department of Computer Science and Engineering\\
    Gothenburg, Sweden}
\begin{document}
%
\maketitle
\thispagestyle{fancy} 
\fancyhf{} 
\renewcommand{\headrulewidth}{0pt} 

\fancyfoot[C]{ 
    \scriptsize  
    \vspace{5pt} 
    
    \parbox{\textwidth}{ 
        \centering
        \textit{Copyright 2026 IEEE. Published in ICASSP 2026 - 2026 IEEE International Conference on Acoustics, Speech and Signal Processing (ICASSP), scheduled for 3-8 May 2026 in Barcelona, Spain. Personal use of this material is permitted. However, permission to reprint/republish this material for advertising or promotional purposes or for creating new collective works for resale or redistribution to servers or lists, or to reuse any copyrighted component of this work in other works, must be obtained from the IEEE. Contact: Manager, Copyrights and Permissions / IEEE Service Center / 445 Hoes Lane / P.O. Box 1331 / Piscataway, NJ 08855-1331, USA. Telephone: + Intl. 908-562-3966.}
    }
}

\begin{abstract}
The objective of domain generalization (DG) is to enable models to be robust against domain shift. DG is crucial for deploying vision-language models (VLMs) in real-world applications, yet most existing methods rely on domain labels that may not be available and often ambiguous. We instead study the DG setting where models must generalize well without access to explicit domain labels. Our key idea is to represent an unseen target domain as a combination of latent domains automatically discovered from training data, enabling the model to adaptively transfer knowledge across domains. To realize this, we perform latent domain clustering on image features and fuse domain-specific text features based on the similarity between the input image and each latent domain. Experiments on four benchmarks show that this strategy yields consistent gains over VLM-based baselines and provides new insights into improving robustness under domain shift.
\end{abstract}
\section{Introduction}
\label{sec:intro}
Traditional machine learning methods often assume training and test data are drawn from the same distribution. However, in practical applications such as autonomous driving and intelligent robotics, models must operate in complex and unpredictable environments, where it is nearly impossible for training data to cover all possible scenarios. This discrepancy, known as domain shift, is the central concern of the Domain Generalization (DG) problem \cite{9847099}, which aims to leverage training data from one or multiple source domains so that the model performs well on unseen target domains.

With the rapid advancement of language models in natural language processing (NLP), vision models have been integrated with them, giving rise to vision-language models (VLMs) \cite{10445007} such as CLIP \cite{radford2021learning}. Pre-trained on large-scale image–text pairs, VLMs exhibit strong zero-shot classification capabilities \cite{radford2021learning}. However, zero-shot classification requires carefully designed prompts, and manually crafting prompts is both time-consuming and suboptimal. To address this, CoOp \cite{zhou2022learning} introduced prompt learning, replacing hand-crafted prompts with learnable parameters (soft prompts) optimized on downstream tasks. While such methods surpass manual prompts in classification, soft prompts are prone to overfitting source domains \cite{ma2023understanding}, thus limiting generalization and yielding unsatisfactory DG performance. This motivates us to improve prompt learning for more robust generalization under domain shifts.

Many high-performing DG methods rely on domain labels during training \cite{bai2024soft, xu2024ensembling, yu2024clipceil}. Yet in practice, assigning precise domain labels to each image is difficult, and domain boundaries are often ambiguous. For example, in autonomous driving, images under slightly different weather or lighting conditions may not fit predefined categories such as “sunny” or “cloudy.” In such cases, algorithms depending on explicit domain labels often fail, limiting their applicability.

Inspired by \cite{matsuura2020domain}, we propose \textbf{Latent Domain Prompt Fusion (LDPF)}, which automatically discovers latent domains and adapts prompts accordingly to enhance the DG ability of VLMs. Our method performs latent domain clustering on domain-specific style features, removing the need for domain labels and potentially capturing intrinsic characteristics more accurately than manual annotations. We design a dual-part soft prompt with domain-agnostic and domain-specific components, balancing invariant and specialized information. Finally, we introduce a domain-similarity–based fusion module that dynamically combines learned soft prompts during inference, improving visual–textual alignment.

Our main contributions are:
\begin{enumerate}
    \item A novel soft prompt learning framework (LDPF) for domain generalization of VLMs, which does not rely on domain labels via latent domain clustering, making it suitable for complex real-world scenarios.
    \item A novel fusion mechanism that represents the target domain as a linear combination of source latent domains, enabling adaptive prompt integration and more robust modality alignment.
    \item Extensive evaluation on four benchmarks, with detailed analyses demonstrating the effectiveness of our approach.
\end{enumerate}

\section{METHODOLOGY}
\label{sec:methodology}
\begin{figure*}[htb]
    \centering
    \begin{minipage}[b]{0.75\linewidth}
        \centering
        \includegraphics[width=\linewidth,keepaspectratio]{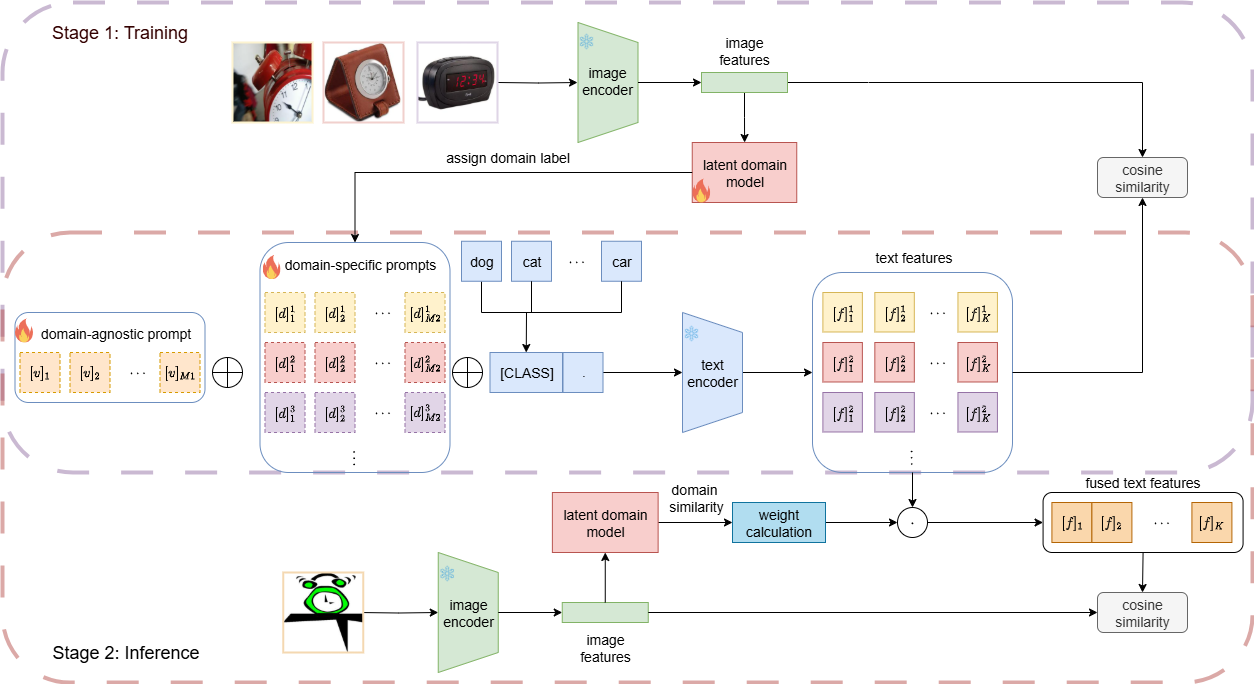}
        \centerline{(a) Latent Domain Prompt Fusion (LDPF) framework.}\medskip
    \end{minipage}%
    \hfill
    \begin{minipage}[b]{0.23\linewidth}
        \centering

        \includegraphics[width=0.8\linewidth,keepaspectratio]{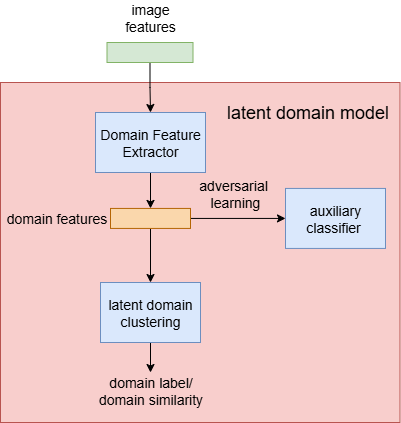}
        \vspace{2mm}
        \centerline{(b) Latent domain model.}
        \medskip 

        \includegraphics[width=0.7\linewidth,height=0.19\textheight,keepaspectratio]{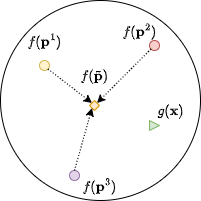}
        \vspace{2mm}
        \centerline{(c) Modality alignment.}
    \end{minipage}

    \caption{Overview of the proposed framework. During training, the image and text encoders are frozen, and the Latent Domain Model assigns latent domain labels to guide the learning of domain-specific prompts. At inference, the model estimates the similarity between the input image and each latent domain, and fuses text features accordingly for better modality alignment, i.e., increasing the cosine similarity between image feature and the corresponding positive text feature.}
    \label{fig:framework}
\end{figure*}

\subsection{Problem setup}
Specifically, we consider a Multi-Source Domain Generalization (MSDG) problem with \( L \) source domains $S_l=\{(x_i^l,y_i^l)\}_{i=1}^{n_l}$, each associated with a joint distribution $P_{XY}^l$. Note that $P_{XY}^l \neq P_{XY}^{l'}, \forall l, l'\in\{1,\cdots,L\}$ and $l\neq l'$. The goal of the MSDG problem is to learn a predictive function $f: \mathcal{X} \to \mathcal{Y}$ using source domain data such that it minimizes the prediction error on an unseen target domain $S_{target}, P_{XY}^{target} \neq P_{XY}^l, \forall l\in \{1,\cdots,L\}$ \cite{9847099}:
\begin{equation}
    \min_f \mathbb{E}[\mathcal{L}(f(x^{target}), y^{target})],
\end{equation}
where $\mathcal{L}(\cdot,\cdot)$ is the loss function. For simplicity, we assume that the source and target domains share the same label set.

\subsection{Soft prompt design}
In our settings, each prompt is divided into three parts: domain-agnostic, domain-specific, and class label:
\begin{equation}
    \mathbf{p}_k=\underbrace{[v]_1[v]_2\cdots[v]_{M1}}_{\text{domain-agnostic tokens}}\:\underbrace{[d]_1[d]_2\cdots[d]_{M2}}_{\text{domain-specific tokens}}[\text{CLASS}]_k.
\end{equation}
Here, \([v]\) represents the domain-agnostic tokens, \([d]\) is the domain-specific tokens, and \([\text{CLASS}]\) represents the class label. \( M_1 \) and \( M_2 \) are hyperparameters that control the lengths of the two types of prompts, and \( k \) represents the \( k \)-th class.

All domains share a single domain-agnostic prompt, while an independent domain-specific prompt is trained for each latent domain. The training is performed in two stages. In the first stage, only the domain-specific prompts are used, and the model minimizes the classification loss $\mathcal{L}_{dsp}$ within each latent domain:
\begin{equation}
    \mathcal{L}_{dsp}=-\frac{1}{N}\sum_{s=1}^{N_s}\sum_{(x,y)\in \mathcal{D}_s}\sum_{k=1}^Ky_k\log P(\hat{y}=k|\mathbf{x, d^s}),
\end{equation}
Here, $N$ denotes the total number of samples, $N_s$ is the number of latent domains, $\mathcal{D}_s$ represents the set of all samples assigned to domain $s$, and $\mathbf{d^s}$ denotes the domain-specific prompt corresponding to domain $s$. 

In the second stage, the domain-agnostic prompt is concatenated with the domain-specific prompts, and the model minimizes the classification loss $\mathcal{L}_{dap}$ across all domains:
\begin{equation}
    \mathcal{L}_{dap}=-\frac{1}{N}\sum_{(x,y)\in\mathcal{D}}\sum_{k=1}^Ky_k\log P(\hat{y}=k|\mathbf{x, p}),
\end{equation}
where $\mathcal{D}$ represents the set of all samples. During this step, only the domain-agnostic prompt is updated, ensuring that it serves as a complementary component that captures domain-invariant knowledge, thereby enhancing generalization.

\subsection{Latent domain model}
To extract domain features from the outputs of the image encoder, we design a two-layer MLP network, referred to as the Domain Feature Extractor. To ensure that the extracted domain features are as independent of class information as possible, we employ an auxiliary classifier trained in an adversarial manner against the extractor:
\begin{equation}
    \mathcal{L}_{adv}=-\frac{1}{N}\sum_{(x,y)\in\mathcal{D}}\sum_{k=1}^Ky_k\log P(\hat{y}=k|h(f(\mathbf{x}))),
\end{equation}
where $h(\cdot)$ represents the auxiliary classifier, and $f(\cdot)$ denotes the image encoder. The resulting domain features are then clustered using k-means, which assigns each sample a latent domain label. We set the number of clusters to $k=N_s$, a hyperparameter tuned via the validation set, strictly independent of ground-truth domain labels. The cluster centroids are stored and later used during inference to compute similarities between input images and latent domains.

\subsection{Prompt fusion mechanism}
This mechanism generates an adaptive text feature that aligns with the target image despite domain shift. We represent the text features from different domains as a set of basis vectors and obtain the fused feature for class $i$ via a weighted linear combination:

\begin{equation}
    \tilde{\mathbf{f}}_k = \sum_{s=1}^{N_s} \alpha_s \mathbf{f}_k^s,
\end{equation}
where $\mathbf{f}_k^s$ is the text feature of class $k$ in domain $s$, and $\alpha_s$ is its fusion weight. The weights are derived from the cosine similarity between the domain feature and each latent domain centroids $\mathbf{c}_s$:

\begin{equation}
    \alpha_s = \frac{\exp(\cos(h(f(\mathbf{x}_i)), \mathbf{c}_s)/\tau)}{\sum_{j=1}^{N_s} \exp(\cos(h(f(\mathbf{x}_i)), \mathbf{c}_j)/\tau)}.
\end{equation}

\subsection{Training and inference}
During training, the backbone encoders are kept frozen, and only the parameters of the soft prompts and the Latent Domain Model are updated. We adopt a Gradient Reversal Layer (GRL) \cite{ganin2015unsupervised} to implement adversarial learning. The Kuhn-Munkres algorithm \cite{munkres1957algorithms} is used to stabilize cluster–label assignment. Total loss function is as follows:
\begin{equation}
    \mathcal{L}=\mathcal{L}_{dsp}+\lambda(\mathcal{L}_{dap}-\mathcal{L}_{adv}),
\end{equation}
where $\lambda=\frac{2}{1+\exp(-10p)}-1 $ \cite{ganin2015unsupervised}. At inference, the Latent Domain Model estimates domain similarities, which guide the adaptive fusion of text features. The classification probabilities that the input image \( \mathbf{x}_i \) belongs to class \( k \) can then be computed using the following formula:

\begin{equation}
    P(y_i=k|\mathbf{x}_i, \tilde{\mathbf{f}})=\frac{\exp(\cos(\tilde{\mathbf{f}}_k, f(\mathbf{x}_i))/\tau)}{\sum_{j=1}^K\exp(\cos(\tilde{\mathbf{f}}_j, f(\mathbf{x}_i))/\tau)}.
\end{equation}
\section{EXPERIMENTS}
\label{sec:experiments}
\subsection{Datasets and implementation details}
We conduct experiments on four benchmark datasets: Office-Home \cite{venkateswara2017deep}, mini-DomainNet \cite{9540778}, PACS \cite{li2017deeper}, and Terra Incognita \cite{beery2018recognition}. Our method builds on CLIP with a ViT-B/16 \cite{dosovitskiy2020image} backbone, where the encoders are kept frozen during training. We set the domain-agnostic prompt length to 4, the domain-specific prompt length to 8, the number of latent domain to 3, while following the same hyperparameter settings as CoOp. Trained for 20 epochs on Terra Incognita, and for 30 epochs on the other datasets. For all datasets, we followed the leave-one-domain-out evaluation protocol as in \cite{9540778}. Each experiment was conducted three times, and we report the average performance along with the standard deviation.

\subsection{Main results}
For comparison, we adopt a series of VLM-based baselines. Besides zero-shot CLIP, we include several prompt-learning approaches, as CoOp, CoCoOp \cite{zhou2022conditional}, BPL \cite{derakhshani2023bayesian}, StyLIP \cite{bose2024stylip}, and DDSPL \cite{xu2024ensembling}, as well as an adapter-based method CLIPCEIL \cite{yu2024clipceil}. It is worth noting that both DDSPL and CLIPCEIL require access to domain labels during training, and we add them as reference.

\begin{table*}[htbp]
    \centering
    \caption{Performance comparison of different methods. We reproduce the results of Zero-shot CLIP and CoOp, while $^*$ indicates that the result is reported in \cite{xu2024ensembling}, $^\dagger$ indicates that the result is reported in \cite{yu2024clipceil}. All methods employed the ViT-B/16 backbone. The best result is highlighted in \textbf{bold} and the second is \underline{underlined}.}
    \label{tab:results}
    \begin{tabular}{llcccccc}
        \toprule
        \textbf{Category} & \textbf{Method} & \textbf{Office-Home} & \textbf{mini-DomainNet} & \textbf{PACS} & \textbf{Terra Inc} & \textbf{Average} \\
        \midrule
        \multirow{6}{*}{Without domain label} 
            & Zero-shot CLIP & 82.03 & 84.10 & 96.10 & 33.95 & 74.05 \\
            & CoOp \cite{zhou2022learning} & 83.52 & 84.42 & 96.32 & 47.52 & 77.92 \\
            & CoCoOp \cite{zhou2022conditional} & 80.70$^*$ & \underline{85.81}$^*$ & 96.73$^*$ & \textbf{50.40}$^\dagger$ & 78.41 \\
            & BPL$^*$ \cite{derakhshani2023bayesian} & 84.02 & 85.28& \underline{96.86} & - & - \\
            & StyLIP$^*$ \cite{bose2024stylip} & \underline{84.63} & - & \textbf{98.05} & - & - \\
            & LDPF (Ours) & \textbf{85.13}\scriptsize{$\pm$0.32} & \textbf{85.82}\scriptsize{$\pm$0.17} & 96.81\scriptsize{$\pm$0.15} & \underline{47.65}\scriptsize{$\pm$0.96} & 78.85 \\
        \midrule
        \multirow{2}{*}{With domain label} 
            & CLIPCEIL$^\dagger$ \cite{yu2024clipceil} & 85.43 & - & 97.55 & 52.98 & - \\
            & DDSPL$^*$ \cite{xu2024ensembling} & 85.59 & 86.23& 98.08& - & - \\
        \bottomrule
    \end{tabular}
\end{table*}

The results are shown in Table \ref{tab:results}. From the results, our method consistently outperforms strong baselines such as Zero-shot CLIP, CoOp, and CoCoOp. In particular, on Office-Home and mini-DomainNet, it achieves the second-best average accuracy, only behind DDSPL, while surpassing all other methods that do not rely on domain labels. On PACS and Terra Incognita, the performance is comparatively weaker, suggesting room for further improvement. We analyzed the possible reasons in Section \ref{sec:upper bound}. Overall, our approach yields an average gain of +4.8\% over Zero-shot CLIP, indicating that it provides a robust generalization boost to CLIP despite not relying on domain labels.

\subsection{Ablation study}
\begin{table}[htbp]
    \centering
    \caption{Ablation study results on Office-Home dataset.}
    \label{tab:ablation}
    \begin{tabular}{lc}
        \toprule
        \textbf{Model} & \textbf{Average} \\
        \midrule
        Remove DAP & 84.91 \\
        Remove DSP & 84.30 \\
        Remove $L_{adv}$ & 84.80 \\
        Remove latent domain clustering & 84.53 \\
        Greedy Fusion & 84.52 \\
        Average Fusion & 85.05\\
        LDPF (Ours) & \textbf{85.13}\\
        \bottomrule
    \end{tabular}
\end{table}

Table \ref{tab:ablation} reports the results of our ablation study. Removing any single component leads to performance degradation, confirming their effectiveness. The first two tests highlight the critical role of domain-specific prompts in our framework. Interestingly, replacing latent domain clustering with manually annotated domain labels results in worse performance, suggesting that human-defined domains may not fully capture image-specific styles and can even mislead the model. Experiments on the fusion mechanism further show its ability to leverage inter-domain similarities for fusion, though the gain over simple averaging remains modest, indicating potential for more advanced fusion strategies in future work.

\subsection{Upper bound analysis}

We train one soft prompt per latent domain and fuse their text features at inference. As a tractable oracle, we consider \textbf{selection} instead of fusion: for each sample, an oracle chooses the most accurate single prompt. This yields the bound
\begin{equation}
    U_{\text{sel}}=P\big(\exists\,s:\,\hat y_s(\mathbf{x})=y\big),
\end{equation}
i.e., the probability that at least one domain-specific prompt predicts correctly. Note that selection is a special case of fusion (a degenerate convex combination with a one-hot weight), and fusion can also create new representations by combining prompts; therefore the true oracle for fusion satisfies $U_{\text{fuse}}\!\ge\! U_{\text{sel}}$. In our analysis, we report $U_{\text{sel}}$ as a conservative, easy-to-compute upper bound.

\label{sec:upper bound}
\begin{figure}
    \begin{minipage}[b]{.48\linewidth}
      \centering
      \centerline{\includegraphics[width=4.0cm]{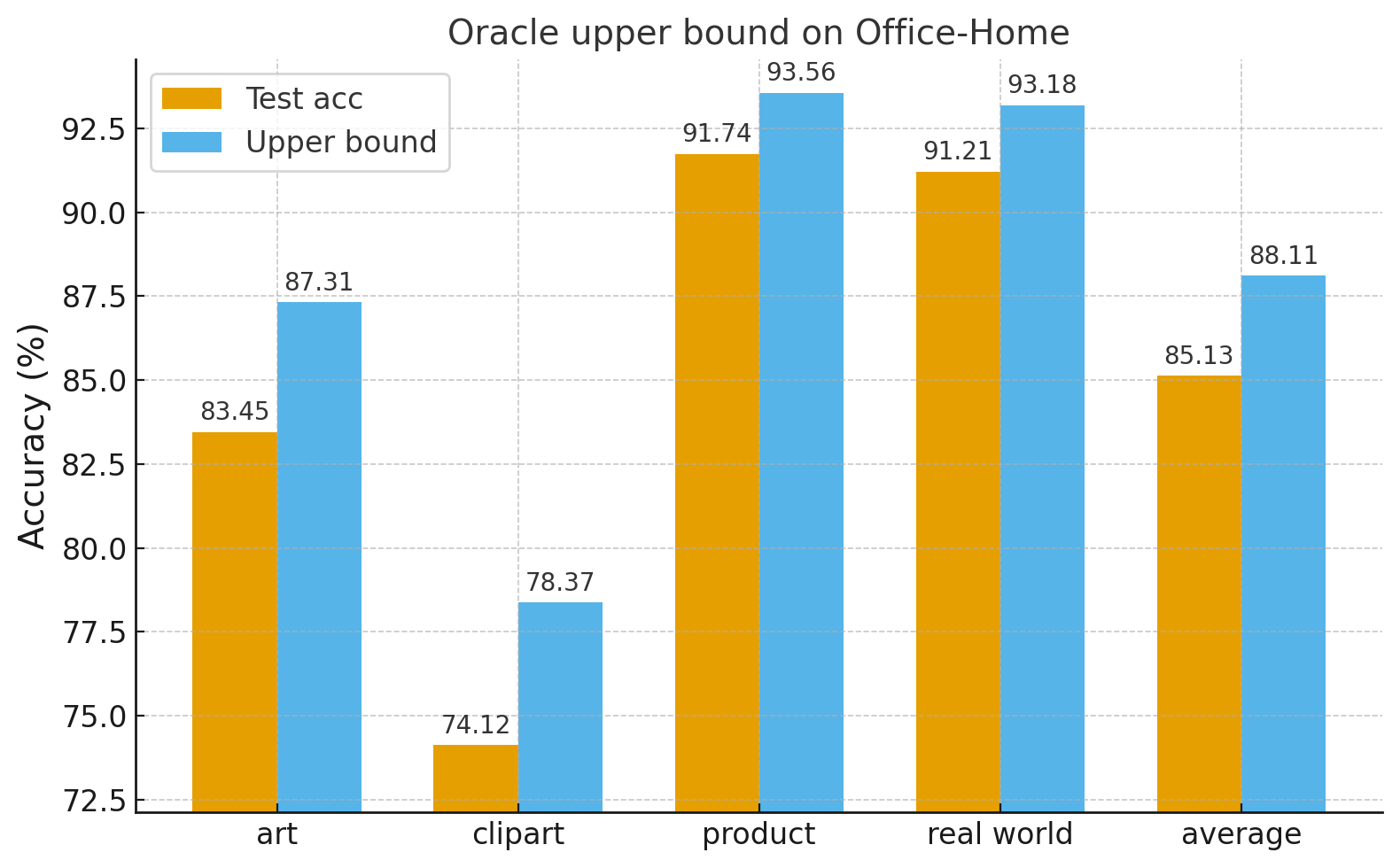}}
      \centerline{(a) Office-Home}\medskip
    \end{minipage}
    \hfill
    \begin{minipage}[b]{0.48\linewidth}
      \centering
      \centerline{\includegraphics[width=4.0cm]{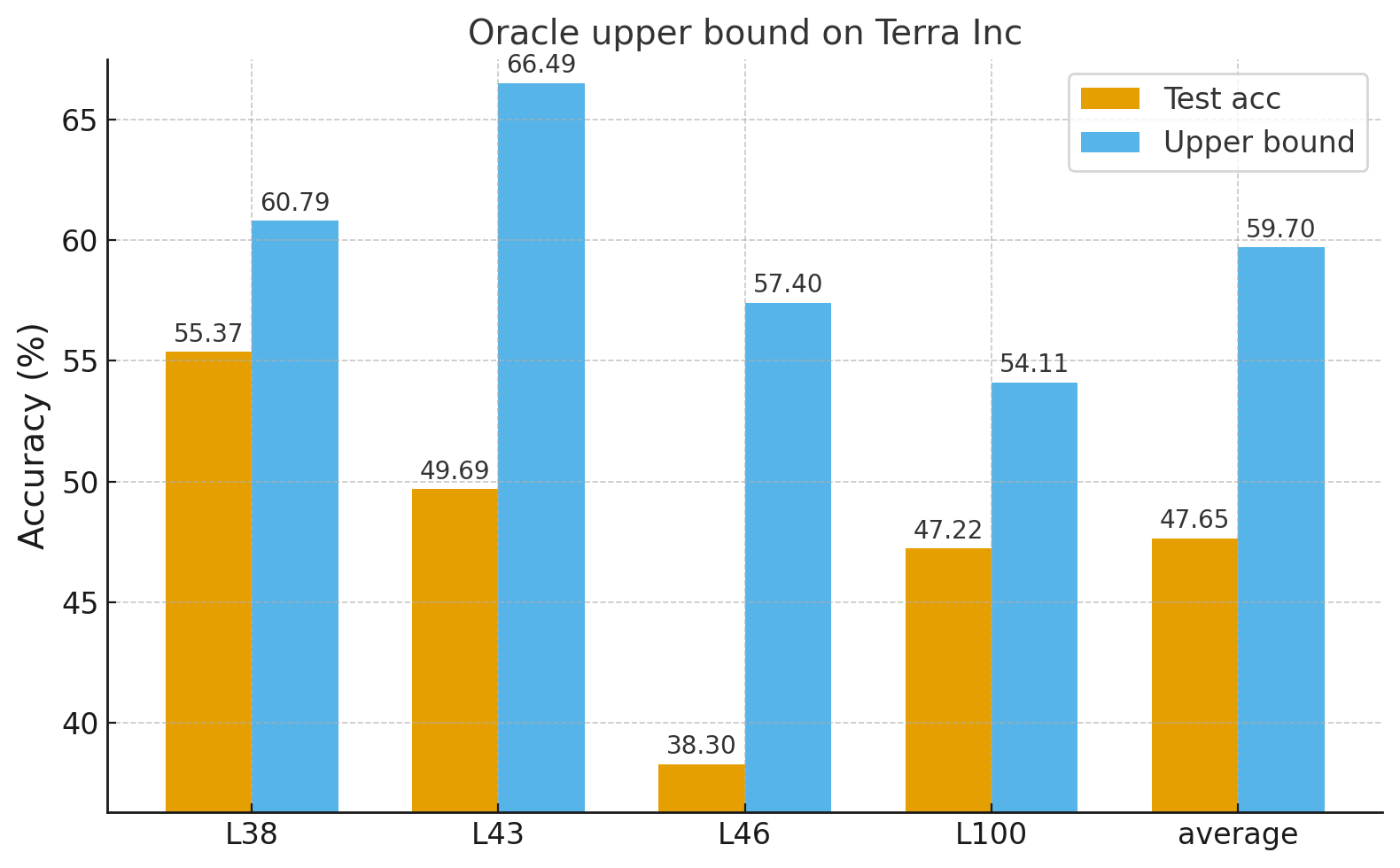}}
      \centerline{(b) Terra Incognita}\medskip
    \end{minipage}
    \caption{Oracle upper bound $U_{\text{sel}}$ on two datasets.}
    \label{fig:upper bound}
\end{figure}

We illustrate the results on the two most representative datasets, Office-Home and Terra Incognita, in Fig. \ref{fig:upper bound}. The other two datasets show trends similar to Office-Home. As observed, on Office-Home the gap between our method and $U_{\text{sel}}$ is only 2.98\%, indicating that the proposed fusion effectively integrates information from different soft prompts to yield accurate predictions. In contrast, on Terra Incognita the gap widens to 12.05\%. This suggests that individual soft prompt tend to be highly specialized: one may handle certain conditions (e.g., infrared/night images) well while others fail badly. A simple similarity-based fusion averages over such inconsistent predictions, often suppressing the correct soft prompt instead of amplifying it. Therefore, Terra Incognita reveals that soft prompt complementarity is strong but cannot be exploited by naive fusion, highlighting the need for more adaptive fusion/selection strategies.

\section{CONCLUSION}
\label{sec:conclusion}

In this work, we studied a prompt learning–based domain generalization method for VLMs without relying on domain labels. Our framework represents an unseen target domain as a linear combination of source latent domains, allowing adaptive transfer of knowledge. Experiments on four benchmarks confirm the effectiveness of this formulation, while our upper bound analysis highlights both the promise and the current limitations of simple fusion. Future work will explore learning-based fusion or gating strategies to better exploit the complementarity among prompts.



\vfill\pagebreak

\bibliographystyle{IEEEbib}
\bibliography{refs}

\end{document}